\documentclass[letterpaper]{article} 
\usepackage{aaai25}  
\usepackage{times}  
\usepackage{helvet}  
\usepackage{courier}  
\usepackage[hyphens]{url}  
\usepackage{graphicx} 
\usepackage{natbib}  
\usepackage{caption} 
\usepackage{algorithm}
\usepackage{algorithmic}
\usepackage{amssymb}
\usepackage{amsmath}
\usepackage{booktabs}
\usepackage{multirow}
\usepackage{pifont}
\usepackage{diagbox}
\usepackage{xcolor}
\usepackage{bm}
\usepackage{newfloat}
\usepackage{listings}
\usepackage{bm}
\DeclareCaptionStyle{ruled}{labelfont=normalfont,labelsep=colon,strut=off} 
\floatstyle{ruled}
\newfloat{listing}{tb}{lst}{}
\floatname{listing}{Listing}
\title{Dynamic Entity-Masked Graph Diffusion Model for histopathological image Representation Learning}
\author{
    Zhenfeng Zhuang\textsuperscript{\rm 1}\equalcontrib,
    Min Cen\textsuperscript{\rm 2}\equalcontrib,
    Yanfeng Li\textsuperscript{\rm 1}\equalcontrib,
    Fangyu Zhou\textsuperscript{\rm 1},
    Lequan Yu\textsuperscript{\rm 3}, 
    Baptiste Magnier\textsuperscript{\rm 4,5}, 
    Liansheng Wang\textsuperscript{\rm 1}\thanks{Corresponding author.}
}
\affiliations{
    \textsuperscript{\rm 1}Department of Computer Science at School of Informatics, Xiamen University, Xiamen, China\\
    \textsuperscript{\rm 2}School of Artificial Intelligence and Data Science, University of Science and Technology of China, Hefei, Anhui, China\\
    \textsuperscript{\rm 3}Department of Statistics and Actuarial Science, The University of Hong Kong, Hong Kong SAR, China\\
    \textsuperscript{\rm 4}EuroMov Digital Health in Motion, Univ Montpellier, IMT Mines Ales, Ales, France\\
    \textsuperscript{\rm 5}Service de Médecine Nucléaire, Centre Hospitalier Universitaire de Nîmes,Université de Montpellier, Nîmes, France\\
    \{zhuangzhenfeng, liyanfeng, lswang\}@stu.xmu.edu.cn, cenmin0127@mail.ustc.edu.cn

}

\usepackage{bibentry}

\begin{document}

\maketitle

\begin{abstract}
Significant disparities between the features of natural images and those inherent to histopathological images make it challenging to directly apply and transfer pre-trained models from natural images to histopathology tasks. Moreover, the frequent lack of annotations in histopathology patch images has driven researchers to explore self-supervised learning methods like mask reconstruction for learning representations from large amounts of unlabeled data. Crucially, previous mask-based efforts in self-supervised learning have often overlooked the spatial interactions among entities, which are essential for constructing accurate representations of pathological entities. To address these challenges, constructing graphs of entities is a promising approach. In addition, the diffusion reconstruction strategy has recently shown superior performance through its random intensity noise addition technique to enhance the robust learned representation. Therefore, we introduce \textbf{H-MGDM}, a novel self-supervised \textbf{H}istopathology image representation learning method through the Dynamic Entity-\textbf{M}asked \textbf{G}raph \textbf{D}iffusion \textbf{M}odel. Specifically, we propose to use complementary subgraphs as latent diffusion conditions and self-supervised targets respectively during pre-training. We note that the graph can embed entities' topological relationships and enhance representation. Dynamic conditions and targets can improve pathological fine reconstruction. Our model has conducted pretraining experiments on three large histopathological datasets. The advanced predictive performance and interpretability of H-MGDM are clearly evaluated on comprehensive downstream tasks such as classification and survival analysis on six datasets. Our code will be publicly available at  https://github.com/centurion-crawler/H-MGDM.
\end{abstract}

%

\section{Introduction}
\begin{figure}[t!]
\includegraphics[width=0.47\textwidth]{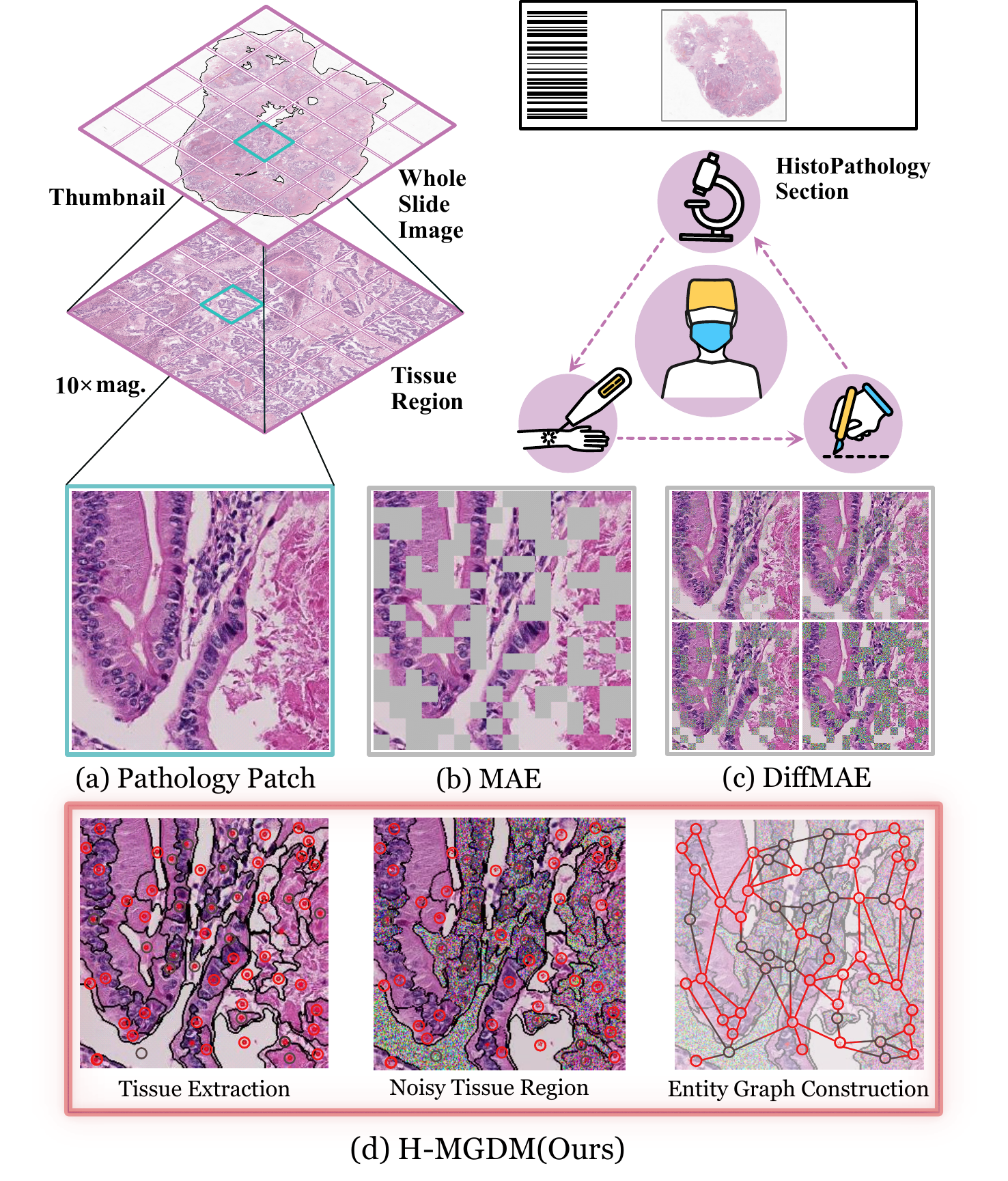}
\centering
\caption{
Pathological slide inspection process from the
overall view to details. Unlike comparison methods, H-MGDM focuses on masked pathological tissue regions rather than grid tiles in patches, constructing the masked subgraph with varying intensities and noise for reconstruction by complementary conditional subgraph.}
\label{fig1}
\end{figure}
Achieving concise and informative histopathology patch image representation is the cornerstone for solving many tasks in the field of computational histopathology analysis, such as cancer diagnosis, grading, segmentation, and prognosis tasks within whole Slide Image (WSI) \cite{panayides2020ai}. 
There are many extensive investigations on pre-training features for pathology images, which can help alleviate the time-consuming and tedious task of manual slide inspection by pathologists \cite{song2023artificial}. 
Today, the representation of pathology image patches relies heavily on transfer learning methods \cite{dosovitskiy2020image,9508542,li2022improving} and the supervised pathology classification models such as KimiaNet \cite{RIASATIAN2021102032}.
The above approaches exist problems like the domain gap and category bias \cite{guan2021domain}, and scarce and high-cost annotations limit retraining. 
Therefore, unlabeled self-supervised learning (SSL) has emerged as a solution to alleviate these limitations by learning salient representations without using labels.

Previous SSL methods like MAE (He et al. 2022) have shown that using masks and reconstruction tasks in SSL effectively enables models to learn from unlabeled pathological data. In the context of pathological images, the topological connections among pathological tissues, including cellular interactions and their surrounding environment, are crucial for various tasks. Recent approaches have underscored the importance of structure function relationships by linking the spatial organization of cells within tissues through cell graphs. These methods enable the extraction of biomarker-based pathological features (Jaume et al. 2021b,a), capturing complex semantic associations that extend beyond pixel-level data to encompass tissues and cells. These approaches align more closely with pathological diagnostic procedures. Entity-based topological analysis provides enhanced control over tissue modeling and facilitates the integration of pathological priors into task-specific histopathological entity representations. This indicates that it is crucial to recognize the interactions among ”Image-to-Graph” based pathological tissue entities. However, when SSL is applied to pathological images, recent studies often focus on using pathological grid tiles in patches as masks, while neglecting the impact of entities (e.g., cellular or tissue regions) mask strategy on the overall semantic representation. Therefore, we introduce a new method to convert images into graphics to capture the structure of pathological entities, such as tissues, in self-supervised learning.

On the other hand, diffusion strategies have recently improved robust learning representations as conditions through their technique of adding noise with varying intensities \cite{purma2023genselfdiff,wei2023diffusion,yang2023diffusion}. 
We propose using an entity-masked graph as the input for the diffusion process, with encoded features from different layers of the graph serving as conditions to maintain strong performance. 
This approach captures powerful and complex entity information, thereby enhancing the representation of pathological images. 

In summary, the motivation of our paper is to better learn the knowledge of entity graphs under self-supervised reconstruction progress. We propose a novel approach that converts histopathological images into entity graphs with dynamic mask and noise for diffusion pre-training to obtain better pathological image representations in the pathology inspection process (see Fig.\ref{fig1}). 
Our contributions are:
\begin{itemize}
\item We propose a novel framework the \textbf{H-MGDM, a novel self-supervised histopathological image representation learning method through Dynamic Entity-Masked Graph
Diffusion Model}. 
A strategy for partially visible entities as conditioning to prompt masked noisy entities to graph diffusion. Random masks and dynamic intensity noises can enhance representations in histopathological images.
\item In H-MGDM, we \textbf{convert pathology images into entity graphs of latent space to incorporate structural information of pathological entities}. 
This allows for more comprehensive spatial and semantic priors.
\item We conducted pretraining experiments on three large histopathological datasets. The advanced predictive performance and interpretability of H-MGDM are clearly evaluated on comprehensive downstream tasks on six datasets. All performance \textbf{across several downstream tasks is averagely improved by 5.18\%}. 
This demonstrates the effectiveness of the H-MGDM framework for pre-training in histopathological image analysis. 
\end{itemize}

\begin{figure*}[t!]
\includegraphics[width=1.0\textwidth]{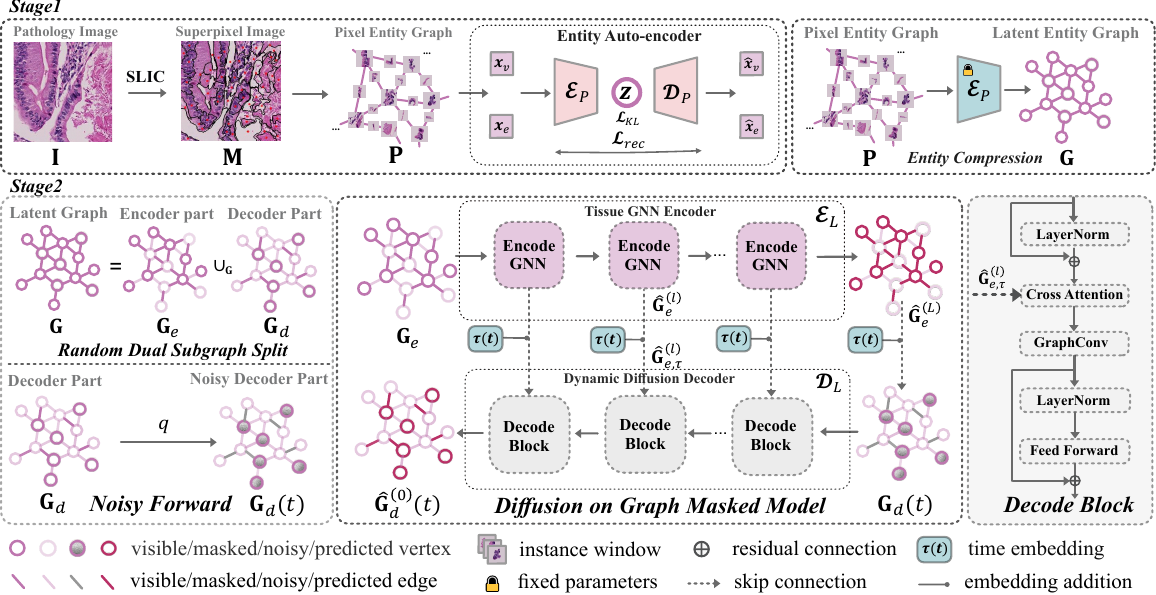}
\centering
\caption{ Overview of the H-MGDM pretraining stages. Conditional diffusion reverse process in the decoder. $\mathbf{G}_e$ and $\mathbf{G}_d$ are two complementary subgraphs of $\mathbf{G}$. $\mathbf{G}_d(t)$ are from the diffusion forward process $q_L$ of $\mathbf{G}_d$. The target is to denoise $\mathbf{G}_d(t)$ to $\hat{\mathbf{G}}_d^{(0)}(t)$ close to $\mathbf{G}_d$ at sampling time $t$. 
} \label{fig2}
\end{figure*}

\section{Related Works}
\subsection{Graph Representation for Digital Pathology}
Recently, graph neural networks (GNN) have been employed to represent patches as graphs for pathology tasks. CGC-Net \cite{zhou2019cgc} introduces a cell-graph convolutional neural network that converts large histology images into graphs, where vertices represent nuclei and edges denote cellular interactions based on vertex similarity. HACT \cite{pati2022hierarchical} develops a hierarchical cell-to-tissue graph representation to jointly model both low-level and high-level graphs, incorporating intra- and inter-level coupling based on the topological distributions and interactions among entities. SHGNN \cite{hou2022spatial} proposes a novel spatial hierarchical GNN framework, equipped with a dynamic structure learning module, to capture entity location attributes and semantic representations, thereby extracting the characteristics of different entities in images. However, these methods focus supervision on global representation. Our proposed method introduces entities subgraph self-supervised targets. This enables entities to capture the contextual implications of local information.

\subsection{Denoising Diffusion Models}
Diffusion models \cite{rombach2022high,ho2020denoising} are known for their ability to generate sophisticated images under conditional control. Diffusion models also exhibit variable masking capabilities and have also been used to enhance representation learning in the self-supervised learning domain, in learning paradigms such as DiffAE \cite{preechakul2021diffusion}. Also, GenSelfDiff-HIS \cite{purma2023genselfdiff} proposes a diffusion-based generative pre-training process for self-supervision to learn efficient histopathological image representations. DiffMAE \cite{wei2023diffusion} integrates diffusion's nuanced detail reconstruction capabilities with MAE's comprehensive semantic representation capability, symbolizing a convergence of methodologies in pursuit of enhanced representation. 

\subsection{Mask for Self-Supervised Representation}
Since the introduction of BERT \cite{devlin2018bert} in language models, masking prediction has reattracted attention. Lots of self-supervised masking methods have been proposed in vision tasks, delineating enhanced representation techniques rooted in various theoretical frameworks: MAE \cite{he2022masked} devised an asymmetric architecture tailored for pixel-level reconstruction, while MixMIM \cite{liu2022mixmim} endeavors to narrow the chasm between pre-training and fine-tuning through the stochastic amalgamation of masks. Additionally, GraphMAE \cite{hou2022graphmae} and GraphMAE2 \cite{hou2023graphmae2} advocate employing masked graph convolution to facilitate feature reconstruction, but the mentioned methods are confined to fixed-intensity masking approaches. The variation mechanism forces the features to adapt to each situation.

\section{Methodology}
Figure \ref{fig2} illustrates the framework of H-MGDM. First, the histopathological entity graph is constructed using the superpixel algorithm SLIC \cite{achanta2012slic} to extract the topological relations among tissues in the image and compress entities to latent space. Then, the GNN encoder is used to encode the visible subgraph as a condition. The latent graph diffusion model is introduced to reconstruct the dynamic self-supervised target of the masked subgraph to obtain robust representations of patches.

\subsection{Pathological Entity Graph Construction}
\label{PEGC}
To strengthen the concept of entity within limited structural constraints, we utilize priori pathological tissue superpixels as tissue entities when constructing the graph. First, a pathological image $\mathbf{I} \in \mathbb{N}^{h_\mathbf{I} \times w_\mathbf{I} \times 3}$ with the height $h_\mathbf{I}$ and the width $w_\mathbf{I}$ is partitioned via SLIC algorithms \cite{achanta2012slic} resulting in a set of superpixels. For each superpixel, $s$, a window of size $a\times a$ centered at $s$ is considered as a vertex $v$ of the pathological entity graph $\mathbf{P}$ in pixel space. Pixels in the window that do not belong to $s$ are assigned the background color. Subsequently, edges will be established between every two vertices with adjacent boundaries, considering the local interactions between adjacent vertices more. The edge $e$ originates from the region after the expansion operation along the boundary between $s^i$ and another neighboring superpixel $s^j$. Thus, the image $\mathbf{I}$ is transformed into a pathological entity graph $\mathbf{P}(\mathbf{V}_\mathbf{P},\mathbf{E}_\mathbf{P},\mathbf{A},\mathbf{D})$, where $\mathbf{V}_\mathbf{P}=\{s_i\}_{i\in[0,N_\mathbf{V})}$, $\mathbf{E}_\mathbf{P}=\{e_j\}_{j\in[0,N_\mathbf{E})}$ are sets of vertices and edges, respectively. And the adjacency matrix is $\mathbf{A}$ and the degree matrix of $\mathbf{A}$ is $\mathbf{D}$. $N_\mathbf{V}$ and $N_\mathbf{E}$ are the numbers of vertices and edges, respectively.

\subsection{Entity Compression into Latent Space}
Our compression model, based on previous work \cite{kingma2013auto,esser2021taming}, utilizes an auto-encoder in stage 1. Given a pathological entity graph $\mathbf{P}$, the encoder $\mathcal{E}_\mathbf{P}$ transforms each entity $\boldsymbol{x} \in (\boldsymbol{X}_v \cup \boldsymbol{X}_e) \subset \mathbb{N}^{a\times a \times 3}$ in $\mathbf{P}$ into a latent representation $\boldsymbol{z} = \mathcal{E}_\mathbf{P}(\boldsymbol{x})$, where $\boldsymbol{z} \in \mathbb{R}^{l\times l \times c}$. The encoder learns an approximate posterior distriution $q(z|x)=\mathcal{N}(z;\mu(x),\sigma^2(x)I)$, with $\mu(x)$ and $\sigma(x)$ being learned mean and standard deviation of $x$. And the decoder $\mathcal{D}_\mathbf{P}$ then reconstructs the image from this latent space, $\hat{\boldsymbol{x}} = \mathcal{D}_\mathbf{P}(\boldsymbol{z}) = \mathcal{D}_\mathbf{P}(\mathcal{E}_\mathbf{P}(\boldsymbol{x}))$. The downsampling factor of the image is $f=a/l$ and we explore various downsampling factors $f$. After the first stage of training is completed, we infer $\mathcal{E}_\mathbf{P} $ to encode all entities into the latent space, resulting in sets $\mathbf{V}_\mathbf{G}$ and $\mathbf{E}_\mathbf{G}$. Those compose the latent space entity graph $\mathbf{G}(\mathbf{V}_\mathbf{G}, \mathbf{E}_\mathbf{G}, \mathbf{A}, \mathbf{D})$.

\subsection{Latent Graph Diffusion Model}
\label{CDG}
The forward process of the latent diffusion model can add noise to the graph entities, describing the degraded sequence caused by Gaussian noise on the latent space, which does not contain much semantics.
Given a well-defined latent diffusion diffusion  \cite{ho2020denoising} forward process $q_L: \{ \mathbf{G}_d(t)\}_{[0,T]}$ with variance time dependence, and a noise schedule $\{\beta(t)\}_{[0,T]}$ where the integer time $t\in[0,T]$, based on Markov chain and diffusion  characteristics\cite{huang2023conditional,wei2023diffusion}, we have:
\begin{equation}
\scriptsize
\left\{
\begin{aligned}
    & q_L(\mathbf{G}_d(t) | \mathbf{G}_d(t-1)) = \: \mathcal{N}(\mathbf{G}_d(t) | \sqrt{1-\beta(t)}\mathbf{G}_d(t-1), \beta(t)\boldsymbol{I}) \\ 
    & q_L(\mathbf{G}_d(t) | \mathbf{G}_d(0)) = \: \mathcal{N}(\mathbf{G}_d(t) | \sqrt{\overline{\alpha}(t)}\mathbf{G}_d(0),(1-\overline{\alpha}(t) )\boldsymbol{I} )
\end{aligned}.
\right.
\end{equation}
$\mathbf{G}_d(t)$ is reparameterized as $\mathbf{G}_d(t)= \sqrt{\overline{\alpha}(t)} \mathbf{G}_d(0)+\sqrt{1-\overline{\alpha}(t)}\epsilon$, noise follows a normal distribution: $\epsilon \sim \mathcal{N}(\boldsymbol{0},\boldsymbol{I})$, where $\alpha(t) = 1 -\beta(t)$, $\overline{\alpha}(t) = \prod_{i=1}^t \alpha(i)$, signal-to-noise ratio $\{\frac{\alpha(t)}{\beta(t)}\}_{[0,T]}$ are chosen to noise gradually that $q_L(\mathbf{G}_d(T)) \approx \mathcal{N}(\boldsymbol{0},\boldsymbol{I})$.
The conditional diffusion reverse above by modeling the reverse distribution $p_L$ which implies masked part conditioned on visible graph $\hat{\mathbf{G}_e}$ from encoder:
\begin{equation}
    p_L(\mathbf{G}_d(t-1)|\mathbf{G}_d(t),\hat{\mathbf{G}}_e) \sim \mathcal{N}(\boldsymbol{0},\boldsymbol{I}).
\end{equation}
Then, a reverse diffusion network $\mathcal{D}_L$ with graph conditioning on $\{\beta(t)\}_{[0,T]}$ is introduced. Considering the graph structure, we can apply the continuous diffusion process to $\mathbf{V}_d(t)$ and $\mathbf{E}_d(t)$ respectively to facilitate the restoration of noisy latent within subgraph pathological entities.
\subsection{Dynamic Diffusion on Masked Graph Model}
\label{H-MGDM}
In stage 2 illustrated in Fig. \ref{fig2}, an asymmetric auto-encoder mode is employed, utilizing GNN layers \cite{kipf2016semi} as the encoder and ViT variants \cite{dosovitskiy2020image} as the decoder. From the LDM perspective, the encoder also provides encoding conditions for the decoder's denoising process. 
For a graph input {\small $G$}, is dynamically randomly divided into two complementary subgraphs  $\mathbf{G}_e(\mathbf{V}_e,\mathbf{E}_e,\mathbf{A}_e,\mathbf{D}_e)$ for encoder and 
$\mathbf{G}_d(\mathbf{V}_d,\mathbf{E}_d,\mathbf{A}_d,\mathbf{D}_d)$ for decoder according to the given masking ratio $r_m=\frac{N_{\mathbf{V}_d}}{N_\mathbf{V}}=\frac{N_{\mathbf{E}_d}} {\mathbf{N}_E}$. The noise-added $\mathbf{G}_d(t)$ serves as the initial input to the decoder. Furthermore, the topological information $\mathbf{A}_*$ $\mathbf{D}_* (*=e, d)$ is maintained during training.

\subsubsection{Tissue GNN Encoder} 
The latent encoder $\mathcal{E}_L$ employs GNN that integrates tissue vertices and edges for $L$ layers. In our pathological graph-based construction, the latent domains of graph vertices and edges are identical. Therefore, the vertex-based message passing can be used to forward edge latent  for $\hat{\mathbf{G}}_e^{(l)}(V_e^{(l)},E_e^{(l)},A_e,D_e)$ in the $l$-{th} layer:
\begin{equation}
    \mathcal{E}_L^{(l)}:
\left\{
\begin{aligned}
    \mathbf{V}_e^{(l+1)}&=\sigma(\tilde{\mathbf{A}}\mathbf{V}_e^{(l)}\mathbf{W}_\mathbf{V}^{(l)})\\
    \mathbf{E}_e^{(l+1)}&=\sigma(\tilde{\mathbf{A}}^*\mathbf{E}_e^{(l)}\mathbf{W}_\mathbf{E}^{(l)})
\end{aligned},
\right.
\end{equation}
where $\tilde{\mathbf{A}}=\mathbf{D}^{-1/2} \mathbf{A} \mathbf{D}^{-1/2}$. $\tilde{\mathbf{A}}$ and $\tilde{\mathbf{A}}^*$ are the normalized symmetric adjacency matrices of the graph $\mathbf{G}_e$ and the dual graph $\mathbf{G}^*_e$, respectively. $\mathbf{V}_e^{(l)}, \mathbf{E}_e^{(l)}$ are inputs to the $l$-th layer, $\mathbf{W}_\mathbf{V}^{(l)}, \mathbf{W}_\mathbf{E}^{(l)}$ are the vertices and edges weight matrices of graph convolution $\mathcal{E}_L^{(l)}$, and $\sigma(\cdot)$ is a non-linear activation.

\subsubsection{Dynamic Diffusion Decoder}
The decoder $\mathcal{D}_L$ utilizes the conditional latent graph diffusion model. 
The noise level $t$ serves as the forward sampling time during pre-training to generate $\mathbf{G}_d(t) $. Similar to the Transformer architecture \cite{vaswani2017attention}, in the $l$-{th}decode block, cross-attention $ CA(\cdot,\cdot) $ utilizes the visible latent $\hat{\mathbf{G}_e}^{(l)}$ from the $ l$-th encoder layers as the conditional control after adding the time embedding $\tau(t)$: $\hat{\mathbf{G}}_{e,\tau}^{(l)} = \hat{\mathbf{G}_e}^{(l)}+\tau(t)$, aiding in denoising $ \hat{\mathbf{G}_d}^{(l)}(t)$ during decoding. 
And the graph convolution of decoder $C_{d}^{(l)}$ is to perform the message passing of the fused latent according to {\small $A_d$} next.   
The Feed Forward $\overline{FF}(\cdot)$ with layer normalized residual block is employed as the final layer within the Decoder Block to induce feature activation, resulting in $\hat{\mathbf{G}_d}^{(l-1)}(t)$. Then, it proceeds to the subsequent blocks to predict $\hat{\mathbf{G}}^{(0)}_d(t)$:
\begin{equation}
\left\{
\begin{aligned}
    &\hat{\mathbf{G}_d}^{(l-1)}(t) = \overline{FF}(C_{dec}(CA(\hat{\mathbf{G}_d}^{(l)}(t),\hat{\mathbf{G}}_{e,\tau}^{(l)}))\\
    & \hat{\mathbf{G}}^{(0)}_d(t) = \mathcal{D}_L(\mathbf{G}_d(t),t,\{ \hat{\mathbf{G}}^{(l)}_e\}_{l \in [0,L)})
\end{aligned}.
\right.
\end{equation}
The skip connection from $\hat{\mathbf{G}}_{e,\tau}^{(l)}$ forms a U-shaped configuration, typically advantageous for graph representation across different levels and noisy latent restoration.
\begin{table*}
\centering
\renewcommand{\arraystretch}{1.15}
\footnotesize
\resizebox{\linewidth}{!}{
\begin{tabular}{ccc|c|cccccc}
\toprule[1.25pt]
\multicolumn{10}{c}{\textbf{Cancer Subtype / Tissue Classification}}    \\ \hline 
\multicolumn{3}{c|}{\textbf{Strategies}}                                                                     & \multirow{2}{*}{\diagbox{\textbf{Methods}}{\textbf{Datasets}}}            & \multicolumn{2}{c}{Komura et al.}                                 & \multicolumn{2}{c}{PANDA} & \multicolumn{2}{c}{IBD}                  \\ \cline{1-3} \cline{5-10} 
\multicolumn{1}{c}{{\footnotesize Graph}} & \multicolumn{1}{c}{{\footnotesize Mask}} & \multicolumn{1}{c|}{{\footnotesize Diffusion}} &                                     & \multicolumn{1}{c}{{\footnotesize ACC$(\%)$}}         & {\footnotesize F1$(\%)$}                              & {\footnotesize ACC$(\%)$}                           &   {\footnotesize F1$(\%)$}  & {\footnotesize ACC$(\%)$}                           &   {\footnotesize F1$(\%)$}         \\ \hline

\ding{55}                               & \ding{55}                            & \ding{55}                               & SimCLR \cite{chen2020simple}  & 69.24±2.12 & 62.45±1.72 & 61.10±2.19 & 58.48±1.25 & 71.44±1.54 & 69.47±2.43 \\
\ding{55}                                & \ding{55}                            & \ding{55}                               & KimiaNet* \cite{RIASATIAN2021102032}   & 72.66±1.69 & 65.93±1.75  & 67.68±1.88  & 55.40±1.19  & 76.42±0.98 & 70.75±1.37 \\
\ding{55}                               & \ding{52}                            & \ding{55}                               & Dino \cite{caron2021emerging}                           &   78.05±1.36    & 70.64±2.01  & 69.92±1.28  & 64.11±1.31 & 82.45±0.82& 75.23±1.41\\
\ding{55}                                & \ding{52}                            & \ding{55}                              & MAE \cite{he2022masked}                                & 77.37±1.51  & 69.44±0.89 & 70.51±1.78 & 62.73±0.89 & 81.06±0.88 & 76.44±1.02 \\
\ding{52}                                & \ding{52}                           & \ding{55}                               & GraphMAE \cite{hou2022graphmae}    & 76.69±2.10   & 67.60±1.87  & 72.22±2.19 & 60.34±2.55 & 75.85±1.25 & 72.78±0.81 \\
\ding{52}                                & \ding{52}                           & \ding{55}                               & GraphMAE2 \cite{hou2023graphmae2}           & 78.86±3.05 & 65.97±1.05 & 72.56±1.80 & 63.49±2.64 & 78.12±1.42 & 72.32±0.79 \\
\ding{55}                               & \ding{55}                            & \ding{52}                               & DiffAE \cite{preechakul2021diffusion}  & 79.11±1.92 & 68.18±1.74 & 71.54±2.12 & 61.62±1.11 & 81.24±1.48 & 74.51±2.08 \\
\ding{55}                                & \ding{52}                            & \ding{52}                               & DiffMAE \cite{wei2023diffusion}   & 78.14±2.10     & 70.23±2.14  & 71.92±1.22 &  65.82±1.82  & 84.58±1.72 & 74.74±2.15\\ \hline
\ding{52}                               & \ding{52}                            & \ding{52}                               & \multicolumn{1}{c|}{\textbf{H-MGDM (Ours)}} & \textbf{82.06±1.17} &   \textbf{72.41±1.36}   &  \textbf{74.51±1.13}  & \textbf{67.32±1.40} & \textbf{86.23±2.31} & \textbf{78.92±1.79} \\ \hline
\multicolumn{3}{c|}{\multirow{3}{*}{\textbf{Ablation study}}} & w/o edge latents & 80.91±1.92 & 70.85±1.58 & 72.39±1.29 & 65.89±1.82 & 84.11±1.72 & 77.41±1.54\\
\multicolumn{3}{c|}{} & w/o skip connection & 79.29±1.27 & 68.43±1.45 & 70.98±1.75 & 62.11±2.16 & 82.35±2.32 & 74.61±2.14 \\ 
\multicolumn{3}{c|}{} & noise intensity fixed  & 79.70±1.14 & 67.14±1.24 & 72.52±1.65 & 63.73±2.31 & 83.58±1.69 & 77.24±2.53\\ \bottomrule[1.25pt]
\end{tabular}
}
\caption{
Comparing classification performance across methods with ablations. The best results are marked in \textbf{bold}. ``*'' indicates label-supervised of the method. The results are reported in: \textit{mean±std} (where \textit{std} stands for standard deviation).}
\label{tab1}
\end{table*}
\begin{figure*}[t!]
\includegraphics[width=0.97\textwidth]{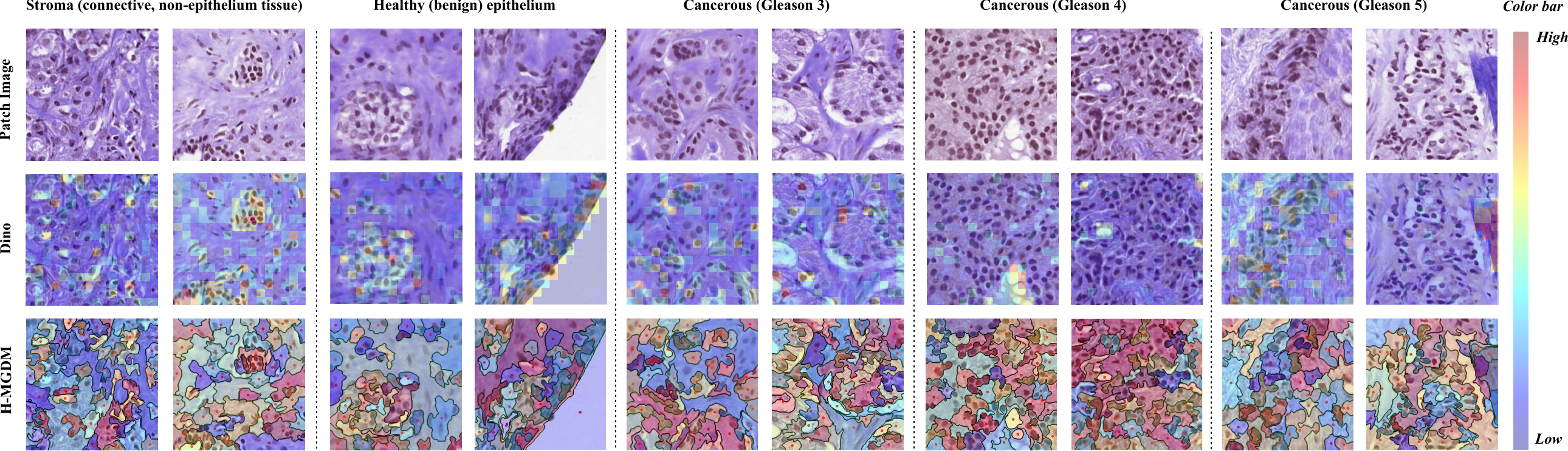}
\centering
\caption{Original images and their attention heatmaps of five different categories of the PANDA dataset, showing the interpretability of our method under the pathological entity graph construction comparison with Dino.}
\label{heatmap}
\end{figure*}
\subsection{Training Strategy}
\subsubsection{Objectives} 
The optimization of the model in pre-taining has two stages: 
In order to avoid high-variance latent space in the first stage, the auto-encoder is optimized by a combination of a reconstruction loss and KL Divergence $D_{KL}$ like VAE \cite{kingma2013auto}. The pixel-space reconstruction constraint $\mathcal{L}_{rec}$ enforces local realism avoids blurriness and ensures that the reconstructions are confined to the image manifold. Here we use the MSE form:
\begin{equation}
\small
\mathcal{L}_{rec} = \mathbb{E}_{q(\boldsymbol{z}|\boldsymbol{x})} \| \boldsymbol{x} - \hat{\boldsymbol{x}}\|_2, 
 ~\mathcal{L}_{VAE} = \mathcal{L}_{rec} + \lambda D_{KL},
\end{equation}
where $\mathbb{E}_{q(\boldsymbol{z}|\boldsymbol{x})}$ represents the expectation of the distribution of the latent variable $z$, $\lambda$ is the loss weight. And $q(\boldsymbol{z}|\boldsymbol{x})$ is the posterior distribution of the latent $\boldsymbol{z}$ given $\boldsymbol{x}$.

The second stage is guided by minimizing the following objective function to optimize parameters $\theta$ of the model. Notably, $\boldsymbol{x}_0$-mode is used dynamic denoising on graph masking of vertices and edges:
\begin{equation}
\small
\begin{aligned}
    \mathcal{L}_{Simple} = &\mathbb{E}_{t,\theta,\epsilon}[\| \mathbf{V}_d(0)- \hat{\mathbf{V}}^{(0)}_d(t) \| + \| \mathbf{E}_d(0)- \hat{\mathbf{E}}^{(0)}_d(t) \|].
\end{aligned}
\end{equation}

Here, $\epsilon$ represents sampled noise. Leveraging variational inference, the well-known Mean Square Error (MSE) objective, derived from the evidence lower bound, is utilized to predict the denoised $\hat{\mathbf{V}}_d^{(0)}$ and $\hat{\mathbf{E}}_d^{(0)}$ as reconstruction targets. 

\subsubsection{Downstream Tasks}
For downstream tasks, we deploy two stages' encoders $\mathcal{E}_\mathbf{P}$ $\mathcal{E}_L$ to inference to obtain the global graph representation $\mathbf{O}_\mathbf{G}$ by readout $r_{\mathbf{G}}$:
\begin{equation}
    \mathbf{G}_o = \hat{\mathbf{G}}^{(L)} = \mathcal{E}_L(\mathcal{E}_\mathbf{P}(\mathbf{G}_\mathbf{P})), \mathbf{O}_\mathbf{G} = r_{\mathbf{G}}(\mathbf{G}_o).
\end{equation}
\begin{itemize}
\item \textbf{Classification.} For the downstream tuning, the cross-entropy loss $\mathcal{L}_{CE}$ is adopted for patch classification tasks. The model is optimized through the classification layer $\text{MLP}_c$ to obtain the predicted $\hat{Y}$ by global representation $\mathbf{O}_\mathbf{G}$ as probabilities of the classes and to supervise it with the classification labels $Y$.

\item \textbf{Regression.} We introduce the Cox proportional hazards model, a semi-parametric regression model. Using survival events and survival time as labels. Here, pathological image features are used to predict risks and analyze the impact on survival. Considering a problem involving two explanatory variables as predictors of survival time $t_i$ and $t_j$ of patients $i$, $j$, $\delta$ represents the termination event (1: death, recurrence, 0: not occurred) $R(t_i)$ represents the condition $t_j > t_i$, for neg log partial likelihood loss:
\begin{equation}
    \mathcal{L}_{Cox}=-\sum_{i:\delta_i=1}[ \hat{h}_i - \log{\sum_{j \in R(t_i)}e^{\hat{h}_j}}],
\end{equation} 
where we use $\text{MLP}_h$ linear map the graph representation ${\mathbf{O}_\mathbf{G}}_i$ to the final hazards prediction $h_i$.
\end{itemize}

\section{Experiments}
\subsection{Experimental Settings}
\subsubsection{Datasets}
\begin{table*}
\centering
\renewcommand{\arraystretch}{1.15}
\footnotesize
\resizebox{\linewidth}{!}{
\begin{tabular}{c|ccccccccc}
\toprule[1.25pt]
\multicolumn{10}{c}{\textbf{Survival Analysis Regression}}    \\ \hline 
 \multirow{2}{*}{\diagbox{\textbf{Methods}}{\textbf{Datasets}}}            & \multicolumn{3}{c}{TCGA-KIRC}                                 & \multicolumn{3}{c}{TCGA-ESCA} & \multicolumn{3}{c}{CRC-PM}                  \\ \cline{2-10}
  & DeepSurv  &  AB-MIL  &  PatchGCN &  DeepSurv & AB-MIL & PatchGCN &  DeepSurv &  AB-MIL  & PatchGCN        \\ \hline
   SimCLR \cite{chen2020simple} & 61.91±4.71 & 62.09±4.33 & 62.46±4.26 & 59.26±4.35 & 58.06±4.60 & 62.59±4.82 & 56.30±2.47 & 58.76±6.51 & 59.87±4.06 \\
                               KimiaNet* \cite{RIASATIAN2021102032}                          & 62.16±4.72 &  64.12±5.28 &  65.76±3.14 &  60.53±6.69 & 59.00±2.97 & 58.16±5.75 & 59.06±9.04 & 59.95±6.11 & 57.67±8.89 \\ 
 Dino \cite{caron2021emerging}  & 67.92±5.61 & 66.94±4.42 & 66.64±3.01    &  57.88±3.91   &  58.85±4.14 & 56.58±5.11 & 58.02±6.72 & 61.01±9.91 & 63.20±8.50 \\
 MAE \cite{he2022masked} & 57.78±2.42 & 61.30±2.54 & 65.08±3.69 & 59.71±5.02 & 57.84±6.29 & 60.62±5.21 & 60.92±7.63 & 59.44±8.98 & 62.80±8.82 \\
 GraphMAE2 \cite{hou2023graphmae2}  & 64.31±6.73 & 66.76±4.21 & 65.84±2.04 & 57.26±5.55 & 59.71±4.90 & 60.35±5.55 & 58.16±6.29 & 59.38±6.94 & 60.42±7.08\\
 DiffAE \cite{preechakul2021diffusion} & 63.26±3.88 & 67.31±1.98  & 68.29±3.84  & 60.74±5.54  &  60.23±5.74 & 63.61±5.49 & 60.81±6.30 & 60.61±5.80 & 62.87±9.52 \\
 DiffMAE \cite{wei2023diffusion}  & 62.41±3.24 & 65.92±5.04 & 67.32±4.18 & 60.60±2.57 & 59.49±4.83 & 63.89±6.32 & 61.16±5.27 & 60.64±6.05 & 61.29±7.12 \\ \hline
 \multicolumn{1}{c|}{\textbf{H-MGDM (Ours)}} & \textbf{66.99±4.56} & \textbf{69.88±3.97} & \textbf{71.17±4.51} & \textbf{62.68±3.04}  &  \textbf{62.55±3.28} & \textbf{64.82±3.45} & 62.27±5.09 & \textbf{63.89±6.94} & \textbf{66.05±7.81} \\ \hline
 w/o edge latent & 65.91±5.74  & 67.88±4.73 & 70.53±5.01 & 59.23±4.29 & 60.62±4.29 & 64.10±4.83 & \textbf{62.32±6.12} & 61.39±7.12 & 65.12±7.28 \\
 w/o skip connection & 63.64±5.32 & 67.00±3.10 & 69.07±5.62 & 58.66±3.85 & 60.34±3.24 & 59.40±2.36 &  
59.78±2.91 & 60.09±4.57 & 60.97±8.36\\ 
 noise intensity $t$ fixed & 63.16±3.62 & 69.44±4.50 & 69.99±4.45 & 59.17±6.49 & 58.97±3.97 & 61.17±3.80 & 60.34±7.38 & 62.05±8.69 & 62.33±5.29 \\ 
 \bottomrule[1.25pt]
\end{tabular}
}
\caption{Survival Analysis performance across SOTAs features on external public validation data by conducting the 5-fold evaluation procedure with 5 runs. The experimental results of CI are reported by \textit{mean±std}.}
\label{tab2}
\end{table*}
\begin{figure*}[t!]
\includegraphics[width=0.95\textwidth]{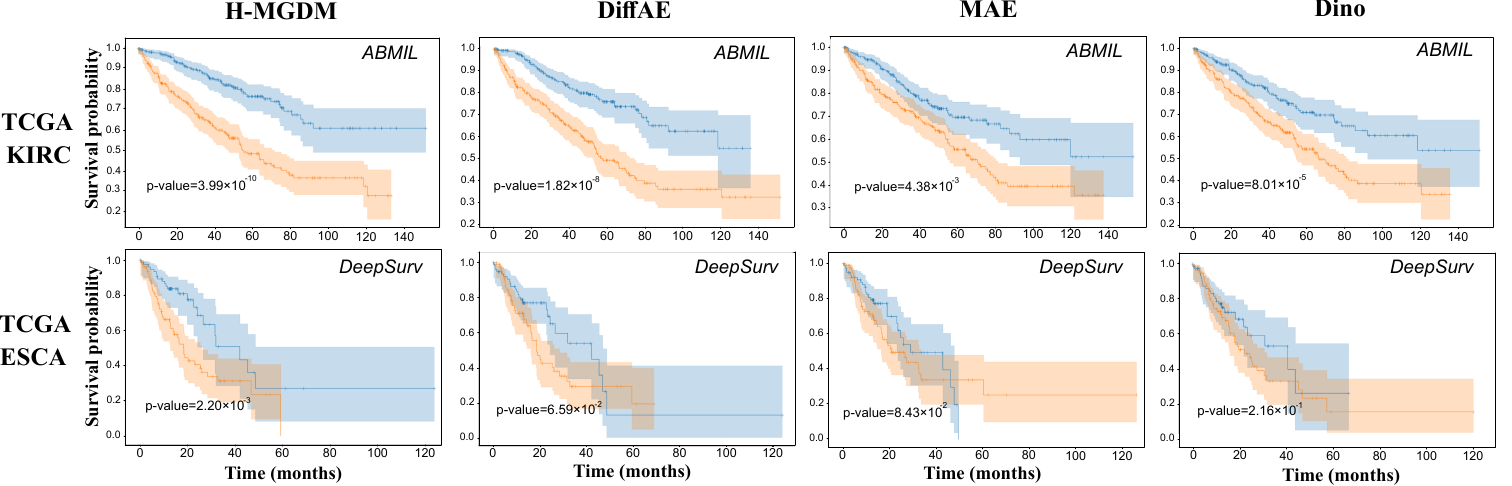} 
\centering
\caption{Kaplan-Meier Analysis of comparison methods and our framework. All patients from the five tests were pooled and analyzed. Each cohort is split into a high-risk (orange) and a low-risk group (blue) according to the median output of the cohort.}
\label{km}
\end{figure*}
The proposed framework underwent pretraining and classification using three large extensive histopathology datasets: \textbf{Komura et al.} \cite{komura2022universal} (1.6M images, 32 cancer types), \textbf{Prostate ANnotation Data Archive (PANDA)} dataset \cite{bulten2022artificial} (11,000 digitized H\&E-stained WSIs, obtaining 12.5M images with 6 level Gleason region annotations), our in-house colorectal cancer data \textbf{IBD} (23M images, with 360K patches annotated into 9 common tissue types). 
For the survival analysis task, we compare the performances on two public cohorts: \textbf{TCGA-KIRC} (512 cases) and \textbf{TCGA-ESCA} (155 cases) and one privately collected primary-metastatic pathology colorectal cancer dataset \textbf{CRC-PM} (388 cases).
We use different methods pre-trained on the pancancer dataset Komura et al. as feature extractors for patches in WSI \footnote{More descriptions in Appendix B.2 about datasets}.
\subsubsection{Implementation~Details}
Experiments involving H-MGDM and comparative methods are conducted with a batch size set to 64 max. SLIC with the initial region number of 500, window size $a$ is 64. Latent downsampling factors $f=2$.
Also, pre-training is optimized by Adam\cite{kingma2014adam} with an initial learning rate of 3e-4 and the plateau scheduler with a minimum learning rate of 1e-5 for 250 epochs. The noise timesteps T is 1000, and the sigmoid schedule $\{\beta(t)\}_{[0, T]}$ from 1e-7 to 2e-3 is used. 
\subsubsection{Evaluation Metrics} 
Accuracy (\textbf{ACC}) and Marco \textbf{F1} are used to evaluate the methods in classification comparison. 
To evaluate the effectiveness of entity latent diffusion, the Root Mean Square Error (\textbf{RMSE}) for graph entities is calculated as a qualitative evaluation metric. Harrell's concordance index (\textbf{C-index} or \textbf{CI}) measures the survival model’s ability to correctly provide a reliable ranking of the survival times based on the individual risk scores. It ranges from 0 to 1, with higher values indicating better performance. 
\subsection{Comparison with Baseline Methods}
\subsubsection{Results}
We conduct comparative experiments among our H-MGDM and other baseline pre-training models: \textit{i.e.} SimCLR \cite{chen2020simple}, Dino, MAE, GraphMAE, GraphMAE2, DiffAE, DiffMAE. We also mark the use of strategies (graph construction, mask strategy, diffusion-guided) by the comparison methods in Table \ref{tab1}. For comparison of the features from various pre-training methods, we use three backbones: DeepSurv \cite{katzman2018deepsurv}, AB-MIL \cite{ilse2018attention} and PatchGCN \cite{chen2021whole} for the survival prediction task. To the best of our knowledge, H-MGDM is the first time these three mechanisms have been introduced simultaneously into histopathology pre-training. Our method achieved outstanding performance due to incorporating structural information of pathological entities and enhancing mask learning during the diffusion procedure. The results are presented in Table \ref{tab1}, \ref{tab2}.
For all baselines, our method achieves an average improvement of 5.99\%, 5.43\%, and 4.146\% on the ACC, F1, and CI. 
\subsubsection{Ablation Study} 
To further explore the proposed components with the effectiveness of our model. We perform ablation experiments in Table \ref{tab1}, \ref{tab2}. The study investigates the influence of edges in graph data, skip connections from encoder representations as diffusion conditioning, and time-varying intensity noise on model performance. The results show these components can enhance the representational capacity of downstream tasks with improvement of 2.50\%, 3.17\%, and  2.46\% on the ACC, F1, and CI metrics. 

\subsubsection{Kaplan-Meier Analysis and Significance}
Based on the median survival risk output by our model, each cancer cohort is divided into high-risk and low-risk groups. If the survival predictions are consistent, the Kaplan-Meier (KM) curves of these groups should show significant differences. As expected, Fig. \ref{km} shows our method's log-rank p-values less than 0.05 for two different cancer cohorts, indicating the effectiveness of in predicting patient survival.
\begin{table}[b!]
\centering
\renewcommand{\arraystretch}{1.15}
\footnotesize
\resizebox{\linewidth}{!}{
\begin{tabular}{c|ccccccccc}
\toprule[1.15pt]
Mask Ratio & 0.1                  & 0.2                  & 0.3                  & 0.4                  & 0.5                  & 0.6                  & 0.7 & 0.8 & 0.9 \\ \hline
Komura et.al.        &  65.87 & 69.90 & 74.44 & 77.07 & 78.38 & \textbf{82.06} & 79.19 & 76.24 & 73.91     \\
PANDA        &    63.12   &   63.15      &   67.88    & 65.37  &  70.51   &  72.49    &  \textbf{74.51}         &     70.47   &  66.24  \\
IBD        &  70.54 &  75.23  &  79.54  &    84.51   &  \textbf{86.23}    &    82.72    &     79.15      &  74.15   &  64.31 \\
\bottomrule[1.15pt]
\end{tabular}
}
\caption{Hyperparameter Investigation of mask ratios}
\label{mask_ratio}
\end{table}
\subsubsection{T-SNE Visualization of pan-cancer representations}
\label{Vis}
\begin{figure}[h!] 
\includegraphics[width=0.48\textwidth]{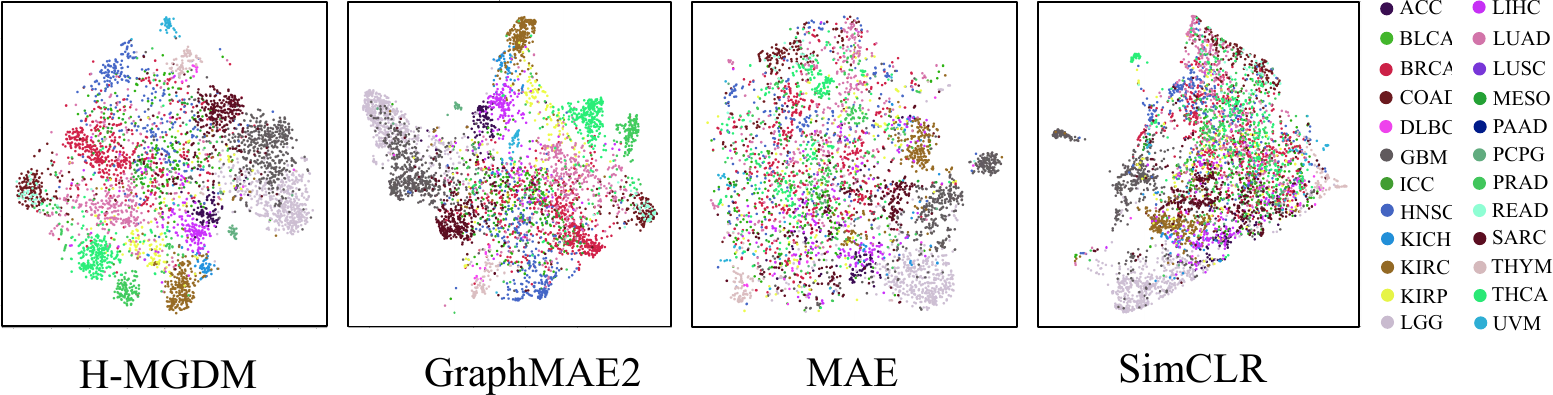} 
\centering
\caption{T-SNE plots of pan-cancer samples' readout representations learning with H-MGDM and baseline methods.}
\label{t-sne}
\end{figure}
 To further confirm the performances of pre-training models, we randomly select 6.4K samples from the trained pre-training model from the Komura et al. \cite{komura2022universal} data test set and visualize the distribution of t-SNE \cite{hinton2002stochastic} as shown in Fig. \ref{t-sne}. Our H-MGDM model can better distinguish 
HNSC (Head and Neck Squamous Cell Carcinoma),
LUAD (Lung Adenocarcinoma) and
UVM (Uveal Melanoma) etc. cancer types. They have high intra-class aggregation and inter-class separation.
\subsection{Analysis of Our Framework}

\subsubsection{Interpretability} The graphical representation of our approach provides scalable interpretability. Unlike posterior explanations of previous methods like Grad-CAM \cite{selvaraju2017grad} and Shap \cite{vstrumbelj2014explaining}, H-MGDM can directly generate high-attention regions during inference readout. We visualize the attention heatmap in Fig. \ref{heatmap} of tissue regions sampling from normal to pathological classes. 
We also visualize the multi-head attention of the Dino method for comparison, highlighting its weaker resistance to interference (some Gleason 4 and 5 samples are focused on blank and outlined regions).
\subsubsection{Masking ratio Investigation}
We examine rates of the dynamic masking strategy, which correspond to the ratio of the target subgraph division. As shown in Table \ref{mask_ratio}, the entity graph masking rate $r_m$ is about 50\%-70\% for reconstruction, and the rest is used as condition learning. Such pre-training features can improve the classification performance of the three datasets. Lower masking rates may hinder the accurate learning of target entity semantics within the visible parts, while higher rates may result in excessive difficulty.
\subsubsection{Decode Strategy Studies}
\begin{table}[h!]
\centering
\renewcommand{\arraystretch}{1.15}
\setlength{\tabcolsep}{1mm}
\footnotesize
\resizebox{\linewidth}{!}{
\begin{tabular}{c|cccccc}
\toprule[1.15pt]
Strategy & NtoN & EtoE & NtoE & EtoN & NtoN \& EtoE & NtoE \& EtoN \\ \hline
ACC     &   73.74   &  71.59   &  71.98   &  68.59  & \textbf{74.51} & 73.82 \\
RMSE     &   0.136   &  0.155   &  0.169   &  0.184  &  0.141  & \textbf{0.127} \\ 
\bottomrule[1.15pt]
\end{tabular}
}
\caption{
{Effectiveness of various decoder conditioning strategies on latent restoration (RMSE) and classification (ACC).}
}
\label{tab4}
\end{table}

The decoder's conditioning strategies need empirical investigation. We test six cross-attention block strategies by aligning different graph attributes between the encoder and decoder (N for vertex, E for edge):  NtoN, EtoE, NtoE, EtoN , NtoN \& EtoE, and NtoE \& EtoN. Graph vertices generally contain more information than edges. PANDA results in Table \ref{tab4} show that graph attribute alignment enhances representation for classification, while attribute heterogeneity improves reconstruction.

\begin{figure}[h!]
\includegraphics[width=0.45\textwidth]{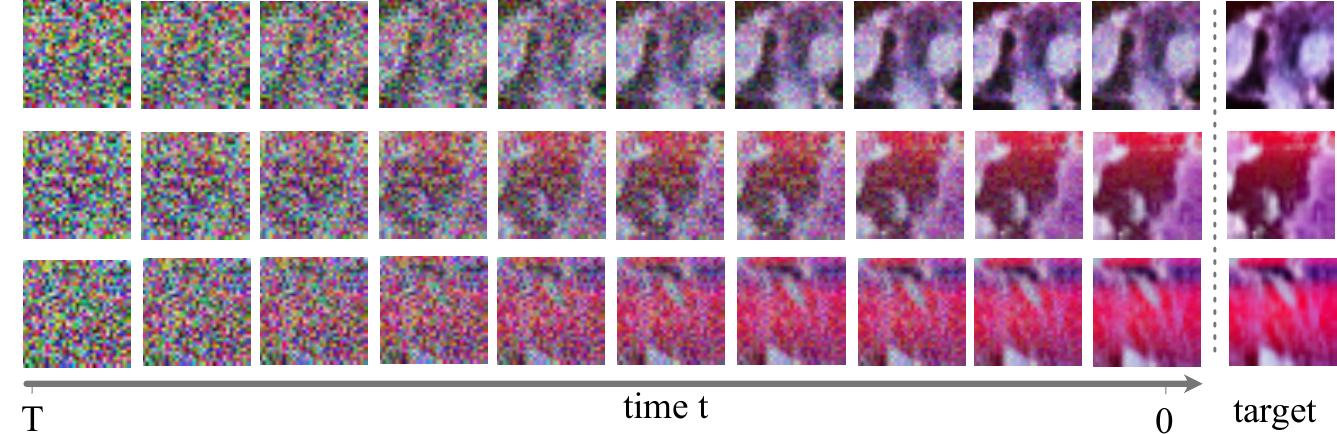} 
\centering
\caption{Visualization of diffusion process over time t.}
\label{fig5}
\end{figure}
\subsubsection{Entity Latent Diffusion Process}
Fig. \ref{fig5} represents the entity restoration as a pre-training proxy task in the latent space. We try to reconstruct fine-grained latent targets throughout the sampling process: as the sampling time $t$ iterates, the latent similarity between diffuse entities and entities increases during the reverse of the process. It Demonstrates excellent representation for latent recovery by the conditioning encoder $\mathcal{E}_L$ in the diffusion decoder $\mathcal{D}_L$ of H-MGDM.

\section{Conclusion}
Our proposed novel framework, the dynamic entity mask on graph diffusion model for histopathology (H-MGDM), addresses the challenge in representation learning and enhances pre-training histopathology representation by incorporating tissue structural information and a suitable masking technique to guide diffusion models during reconstruction. The results of our experiments on six histopathology datasets covering common cancer types demonstrate the superior classification and regression performance of H-MGDM compared to existing methodologies. We anticipate the widespread applicability of H-MGDM in diverse downstream tasks such as prognosis and generation. Additionally, we are committed to exploring the potential of the newer entity extraction methods, the deeper interpretability, and the larger-scale experiments of our framework in our future research endeavors.

\section{Acknowledgments}
This work was supported by National Natural Science Foundation of China (Grant No. 62371409) and Fujian Provincial Natural Science Foundation of China (Grant No. 2023J01005).



\bibliography{aaai25}
\def\maketitlesupplementary
{
\newpage
   \twocolumn[
    \centering
    \Large
    \vspace{0.5em}\textbf{Supplementary Material} \\
    \vspace{1.0em}
   ] 
}
\clearpage
\setcounter{page}{1}
\maketitlesupplementary
\appendix

We firstly provide the H-MGDM inference algorithm and introduce downstream tasks in Appendix A. The comprehensive introduction to the comparison methods is in Appendix B. We also provide additional interpretability and Kaplan-Meier analysis results in Appendix C. The limitations of our approach and areas for future work are shown in Appendix D. 
\section{A. Preliminaries and Details of Inference}
In this part, we make more additions to the graph latent space diffusion model, and model inference for downstream tasks.
\subsection{A.1 Preliminaries of Latent Diffusion on Graph}
Given a latent space instance graph $\mathbf{G}(\mathbf{V}, \mathbf{E}, \mathbf{A}, \mathbf{D})$. DDPM algorithm can generate graphs refer to \cite{liu2023generative}: 
\begin{equation}
\begin{aligned}
    q(\mathbf{G}(t)|\mathbf{G}(t-1)) &= (\mathbf{V}(t-1)\mathbf{Q}_t^\mathbf{V}, \mathbf{E}(t-1)\mathbf{Q}(t)^\mathbf{E}) \\
    q(\mathbf{G}(t)|\mathbf{G}) &= (\mathbf{V} \tilde{\mathbf{Q}}(t)^{\mathbf{V}},\mathbf{E} \tilde{\mathbf{Q}}(t)^{\mathbf{E}})
\end{aligned}
\end{equation}
where $\mathbf{G}(t)=(\mathbf{V}(t),\mathbf{E}(t))$ refers to the noisy graph composed of the node feature matrix $\mathbf{V}_t$ and the edge attribute tensor $\mathbf{E}_t$ at step $t$. $\mathbf{Q}_t^{\mathbf{V}}$ and $\mathbf{Q}_t^{\mathbf{E}}$ refer to the noise added to the node and edge, respectively. This Markov formulation allows sampling directly at an arbitrary time step without computing the previous steps.
And $\mathbf{G}_t$ can denoise with the conditional graph $\mathbf{G}_c$ as follows: 
\begin{equation}
\begin{aligned}
    &p(\mathbf{G}(t-1)|\mathbf{G}(t),\mathbf{G}_c,t) \\
    &= (\mathbf{V}_{t}\mathbf{Q}_t^{CA(\mathbf{V},\mathbf{V}_c)},\mathbf{E}_{t}\mathbf{Q}_t^{CA(\mathbf{E},\mathbf{E}_c)})
\end{aligned}
\end{equation}
we introduce a domain-specific encoder to project $G_c$ into an intermediate representation which is then mapped to the intermediate layers of the UNet structure via a cross-attention $CA(\cdot,\cdot)$ $softmax(\frac{Q_* K_*^T}{V_*}) $ with :
\begin{equation}
\left\{
\begin{aligned}
Q_\mathbf{V} =& W_{Q_\mathbf{V}} \cdot \mathbf{V}, K_\mathbf{V} = W_{K_\mathbf{V}} \cdot \mathbf{V}_c, V_\mathbf{V} = W_{V_\mathbf{V}} \cdot \mathbf{V}_c \\
Q_\mathbf{E} =& W_{Q_\mathbf{E}} \cdot \mathbf{E}, K_\mathbf{E} = W_{K_\mathbf{E}} \cdot \mathbf{E}_c, \mathbf{V}_E = W_{V_\mathbf{E}} \cdot \mathbf{E}_c
\end{aligned}
\right.
,
\end{equation}
here, $W_{Q_\mathbf{V}}$,$W_{K_\mathbf{V}}$,$W_{V_\mathbf{V}}$,$W_{Q_\mathbf{E}}$,$W_{K_\mathbf{E}}$ and $W_{V_\mathbf{E}}$ are learnable projection matrices. Then we learn the conditional LDM on the graph optimized with:
\begin{equation}
    \mathcal{L} = \mathbb{E}_{\mathcal{E}(\mathbf{G}),\mathbf{G}_c,\epsilon \sim \mathcal{N}(0,1),t } [ \| \epsilon - \epsilon_\theta(\boldsymbol{z}_t,t,\tau_\theta(\mathbf{G}_c) \| ],
\label{L_LDM}
\end{equation}
where both $\tau_\theta$ and $\epsilon_\theta$ are jointly optimized via Eq.\ref{L_LDM}. This conditioning mechanism is flexible with the same domain experts. In our method, $\mathbf{G}_c$ are conditioning prompts in the dynamic diffusion decoder.  
\subsection{A.2 Detailed description of Inference and Downstream tasks}
\label{foot1}
\subsubsection{Inference description}
\begin{algorithm}[h!]
\caption{Inference of H-MGDM framework}
\label{alg:infer}
\begin{algorithmic}[1]
\REQUIRE Pathology Image $\mathbf{I}$; Optimized entity compressed encoder $\mathcal{E}_P$ and tissue GNN encoder $\mathcal{E}_L$ with $L$ layers graph convolution, the Readout layer $r_{\mathbf{G}}$;
\ENSURE Pooled and task-specific readout global graph representation $\mathbf{O}_\mathbf{G}$;\\
\textcolor[rgb]{0.7,0.7,0.7}{\textit{\# stage one: entity compression}}
\STATE Superpixel Image $\mathbf{M} \leftarrow \mathbf{I}$ using SLIC;
\STATE Adjacency algorithm constructs tissue entity graph $\mathbf{P}$ from $\mathbf{M}$ with vertex tiles $\mathbf{V}_\mathbf{P}$, edge tiles $\mathbf{E}_\mathbf{P}$, Adjacency matrix $\mathbf{A}$ and Degree matrix $\mathbf{D}$;
\FOR{entity $\boldsymbol{x}_\mathbf{V}$ in $\mathbf{V}_\mathbf{P}$ }
\STATE    $\boldsymbol{z}_\mathbf{V}\leftarrow \mathcal{E}_P(\boldsymbol{x}_\mathbf{V})$
\ENDFOR
\FOR{entity $\boldsymbol{x}_\mathbf{E}$ in $\mathbf{E}_\mathbf{P}$ }
\STATE    $\boldsymbol{z}_\mathbf{E}\leftarrow \mathcal{E}_P(\boldsymbol{x}_\mathbf{E})$
\ENDFOR
\STATE Construct graph $\mathbf{G}$'s vertex and edge sets in latent:
 $\mathbf{V}_{\mathbf{G}} \leftarrow \{\boldsymbol{z}_\mathbf{V}\}$; $\mathbf{E}_{\mathbf{G}} \leftarrow \{\boldsymbol{z}_\mathbf{E}\}$;
\STATE Constructing latent graph $\mathbf{G}(\mathbf{V}_{\mathbf{G}},\mathbf{E}_{\mathbf{G}},\mathbf{A},\mathbf{D})$; \\ \textcolor[rgb]{0.7,0.7,0.7}{\textit{\# stage two: graph encoding}}
\FOR{$i$-th graph convolution $C_{enc}^{(l)}$ in $\mathcal{E}_L$}
    \IF{$i \neq 0$}
        \STATE $\hat{\mathbf{G}}^{(l)} \leftarrow C_{enc}^{(l)}(\hat{\mathbf{G}}^{(l-1)})$
    \ELSE
        \STATE $\hat{\mathbf{G}}^{(1)} \leftarrow C_{enc}^{(1)}(\mathbf{G})$
    \ENDIF
\ENDFOR
\STATE $\mathbf{G}_o = \hat{\mathbf{G}}_e^{(L)} \leftarrow \mathcal{E}_P(\mathbf{G}_{\mathbf{P}})$
\STATE Readout representation $\mathbf{O}_\mathbf{G} \leftarrow r_{\mathbf{G}}(\mathbf{G}_o) $
\end{algorithmic}
\end{algorithm}
As mentioned in Section \textbf{Training Strategy}, the original image first undergoes two pre-training stages of encoders' inference to obtain the readout graph representation $\mathbf{G}_o$ as the final representation of the image $\mathbf{I}$ in algorithm \ref{alg:infer}. In Graph Neural Networks (GNNs), the readout operation creates a global representation of the graph from node or edge embeddings. That includes \textit{Mean Pooling}, \textit{Max Pooling}, \textit{Sum Pooling} and \textit{Attention-based Readout}, etc. This is essential for tasks like graph classification and regression, as it combines local information into a global context, allowing the model to understand the entire graph. And then we use the final global $\mathbf{O}_\mathbf{G}$ to perform downstream tasks.
\subsubsection{Downstream tasks}
\begin{itemize}
    \item Histopathological image classification involves the analysis of microscopic images of tissue samples to identify and classify various diseases, including cancer and other medical conditions. We use the global representation $\mathbf{O}_\mathbf{G}$ obtained from the model and optimize the classification layer $\text{MLP}_c$ to predict the probabilities of the classes and to supervise it with the true labels. The criterion cross-entropy loss is defined as:
    \begin{equation}
    \small
    \mathcal{L}_{CE} = -\frac{1}{B} \sum_{i=1}^{B}\sum_{j=1}^{C} y_{ij}log(\hat{y}_{ij})
    \end{equation}
    where $Y=\{y_{ij}\}$,$\hat{Y}=\{\hat{y}_{ij}\}$,$\hat{y_i} = \text{MLP}_c(\mathbf{O}_\mathbf{G})$, $B$ is the number of one batch, $C$ is the number of classes. Y is the one-hot label matrix for optimization. 
    \item The Cox proportional hazards (PH) regression model is a class of survival models in statistics. Survival models relate the time that passes, before some event occurs, to one or more covariates that may be associated with that quantity of time. In a proportional hazards model, the unique effect of a unit increase in a covariate is multiplicative with respect to the hazard rate. As consisting of two parts: (1) the underlying baseline hazard function, often denoted $\lambda_0(t)$ describing how the risk of event per time unit changes over time at baseline levels of covariates; (2) and the effect parameters, describing how the hazard varies in response to explanatory covariates.
    \item Kaplan-Meier survival curves and the Log-Rank test. To estimate survival probability at a given time, the Kaplan-Meier method utilizes the risk set, incorporating data on censored individuals up to the point of censorship rather than discarding it.
\end{itemize}
\section{B. Introduction of Comparison}
\subsection{B.1 WSI and Patch Preprocessing}
In this part, we follow CLAM \cite{lu2021data} to conduct WSI processing with three stages: segmentation, patching, and label allocation for patches.
\subsection{Segmentation}
The pipeline begins with the automated segmentation of foreground tissue for each digitized slide. The whole slide image (WSI) is initially read at a reduced resolution and converted from the RGB to the HSV color space. A binary mask delineating tissue areas is generated by applying a threshold to the saturation channel after performing median blurring, followed by morphological closing to bridge any gaps. Contours detected in this process are filtered based on their area and then stored for subsequent analysis. The segmentation mask can be visually reviewed, and a text file is produced that lists the processed slides and key parameters used during segmentation.
\subsubsection{Patching} 
After segmentation, the algorithm systematically extracts patches of size
$h_\mathbf{I}\times w_\mathbf{I}$ from the identified tissue regions at the desired magnification. These patches, along with their spatial coordinates and related slide metadata, are saved in the HDF5 hierarchical data format. Depending on the dimensions of the WSI and the selected magnification, the number of patches per slide can range from a few hundred to several hundred thousand.
\subsubsection{Allocation}
The pathology images derived from patching are either sent to pathologists for annotation or are assigned task-specific labels based on pre-existing annotations of the WSI. A script is employed to select homogeneous tissue areas, with an emphasis on avoiding non-structured regions as much as possible.
\subsection{B.2 More descriptions about datasets}
The proposed framework undergoes pretraining and classification using three large extensive histopathology datasets. For the survival analysis task, we used the pre-trained weights from the pan-cancer dataset Komura et al. and compared the performances of different models on two public TCGA cohorts and one privately collected primary-metastatic pathology colorectal dataset. 
\subsubsection{Komura et al.} \cite{komura2022universal} comprises 1.6M image patches of H\&E stained histological samples representing 32 solid cancer types sourced from the GDC legacy database. These images, extracted from 8,736 diagnostic slides of 7,175 patients, were randomly cropped at 6 magnification levels, the size ranging from 128$\times$128 to 256$\times$256 $\mu m$, within annotated regions by trained pathologists \cite{komura2022universal}. 
\subsubsection{Prostate cANcer graDe Assessment (PANDA)} dataset \cite{bulten2022artificial} includes approximately 11,000 whole-slide images of digitized H\&E-stained biopsies. This dataset incorporates annotations for five grades of Gleason scores \cite{humphrey2004gleason} provided by the Radboud University Medical Center and the Karolinska Institute. We sampled patches according to Radboud's Gleason area annotation. Valid annotation values are 0: background (nontissue) or unknown; 1: stroma (connective tissue, non-epithelium tissue); 2: healthy (benign) epithelium; 3: cancerous epithelium (Gleason 3); 4: cancerous epithelium (Gleason 4); 5: cancerous epithelium (Gleason 5). After processing, 12.5M images are generated.
\subsubsection{IBD}
On the privately collected pathological colorectal cancer data set of 23M patches at 10x magnification. And 360K patches are annotated for 9 common tissue types(0: background; 1:tumor; 2:necrotic; 3:mucous; 4:fat; 5:smooth muscle; 6:thrombus; 7:neplasm; 8:normal colon mucosa) for downstream classification tasks. Note that non-pathological section patches will be removed. All data are pre-trained with data augmentation such as random scaling and rotation.
\subsubsection{TCGA-KIRC} The Cancer Genome Atlas Kidney Renal Clear Cell Carcinoma (TCGA-KIRC, 512 histopathology cases) data collection is part of a larger effort to build a research community focused on connecting cancer phenotypes to genotypes by providing clinical images matched to subjects from The Cancer Genome Atlas (TCGA). 
\subsubsection{TCGA-ESCA} The Cancer Genome Atlas Esophageal Carcinoma (TCGA-ESCA, 155 histopathology cases) is part of the TCGA project and is dedicated to studying the genomic characteristics of esophageal cancer.
\subsubsection{CRC-PM} CRC-PM is a private dataset consisting of three batches of pathological images and survival information of rectal cancer and colon cancer (388 histopathology cases), with a censorship rate of $\ge$80\%.  
~\\
All pathological sections WSI are normalized at 10×, and divided into patches. Features are extracted from different pre-trained models for survival analysis experiments.
\subsection{B.3 Baseline Methods}
\subsubsection{SimCLR}

SimCLR (A Simple Framework for Contrastive Learning of Visual Representations) is an unsupervised method that applies random augmentations to each image to create two views encoded into feature vectors by a neural network. These vectors are then transformed into projection vectors. The Normalized Temperature-scaled Cross Entropy (NT-Xent) loss is used to bring similar images closer and separate dissimilar ones. For a positive pair $(i,j)$, the loss function is:
\begin{equation}
    l_{i,j} = -log \frac{\text{exp}(sim(\boldsymbol{z}_i,\boldsymbol{z}_j)/\tau)}{\sum_{k=1}^{2N} \mathbb{1}_{[k\neq i]}\text{exp}(sim(\boldsymbol{z}_i,\boldsymbol{z}_k)/\tau)}.
\end{equation}
where $sim(\boldsymbol{z}_i,\boldsymbol{z}_j)$ is the cosine similarity, $\tau$ is the temperature parameter, $N$ is the batch size, and $\mathbb{1}_[k\neq i]$ ensures the positive pair is excluded from the denominator. 
\subsubsection{KimiaNet}
KimiaNet utilizes the DenseNet-121 architecture with four dense blocks, which have been fine-tuned and trained using histopathology images across various configurations.
\subsubsection{Dino}
DINO is a label-free knowledge distillation method where the student network learns from the teacher's output using cross-entropy loss $\mathcal{L}_{CE}$. The teacher's parameters $\theta_t$ are updated using the student's parameters $\theta_s$ with momentum:
\begin{equation}
    \theta_t \leftarrow m \theta_t + (1-m) \theta_s.
\end{equation}
To avoid collapse, the method includes prediction centering and sharpening of image $\mathbf{I}$:
\begin{equation}
    \tilde{p}_t = p_t -c, p_t(\mathbf{I}) = \frac{p_t(i)^{1/ \tau}}{\sum p_t^{1/\tau}}. 
\end{equation}
DINO works with both CNNs and ViTs without changing the original architecture.
\subsubsection{MAE}
MAE (Masked Autoencoder) is a simple, effective, and scalable approach for visual representation learning. MAE applies the concept of masked autoencoding, inspired by the success of BERT in NLP, where random patches of an image are masked and then reconstructed in pixel space. The method features an asymmetric encoder-decoder design, where the encoder processes only a subset of visible patches, and a lightweight decoder reconstructs the input from the latent representation and mask tokens. This design allows for a high masking rate, significantly reducing redundancy and computational costs, while still learning robust representations.
\subsubsection{GraphMAE}
GraphMAE is a graph neural network model based on an autoencoder. It masks node features during encoding and uses a GNN to create embeddings. In decoding, it remasks the same nodes to improve feature learning. The model learns meaningful information through unsupervised feature reconstruction. The node feature $\tilde{\boldsymbol{x}}_i$ for $\mathbf{v}_i\in \mathcal{V}$ in the masked feature matrix $\tilde{\boldsymbol{x}}$ can be defined as: 
\begin{equation}
\tilde{\boldsymbol{x}}_i = 
\left\{
\begin{array}{ll}
\boldsymbol{x}_{[M]} & \mathbf{v}_i \in \tilde{\mathcal{V}}, \\
\boldsymbol{x}_i & \mathbf{v}_i \notin \tilde{\mathcal{V}}.
\end{array}
\right.
\end{equation}
The objective of GraphMAE is to reconstruct the masked features of nodes in the subset $\mathcal{V}$ given the partially observed node signals $\tilde{X}$ and the input adjacency matrix.
\subsubsection{GraphMAE2}
GraphMAE2 addresses inaccuracies in node feature semantics by enhancing the masked prediction approach. It introduces regularization in the decoding stage by repeatedly re-masking the encoded representations, forcing the decoder to reconstruct features from corrupted data. Additionally, it predicts masked node representations in an embedding space distinct from the input feature space to reduce direct input influence.
It employs the scaled cosine error to measure the reconstruction error and sum over the errors of the multi views $K$ for training:
\begin{equation}
    \mathcal{L}_{input} = \frac{1}{|\tilde{\mathcal{V}}|} \sum_{j=1}^K \sum_{\mathbf{v}_i \in \tilde{\mathcal{V}}} (1- \frac{\boldsymbol{x}_i^T \boldsymbol{z}_i^{(j)}}{\| 
\boldsymbol{x}_i \| \cdot \|\boldsymbol{z}_i^{(j)} \|})^{\gamma}.
\end{equation}
where $\boldsymbol{z}_i^{(j)}$ is the $i$-th row of predicted feature, and $\gamma>=1$  is the scaled coefficient. It also learns the parameters $\theta$ of the encoder and projector by minimizing the following scaled cosine error with gradient descent:
\begin{equation}
    \mathcal{L}_{latent} = (1- \frac{\tilde{\boldsymbol{z}}_i^T \bar{\boldsymbol{x}_i}}{\| \bar{\boldsymbol{z}}\|\cdot \| \bar{\boldsymbol{x}} \|})^{\gamma}. 
\end{equation}
And the parameters of the target generator $\zeta$ are updated via an exponential moving average of $\theta$ using weight decay $\tau$: $\zeta = \tau \zeta + (1-\tau)\theta$. This approach prevents overfitting to input features and enhances the model's generalization capabilities.
\subsubsection{DiffAE}
DiffAE is a novel approach that leverages diffusion probabilistic models (DPM) for image attribute editing by combining the strengths of DPMs with autoencoder architectures. The key idea behind DiffAE is to use DPMs as decoders within an autoencoder framework, where the autoencoder’s encoder outputs a semantically meaningful latent code, denoted as $\boldsymbol{z}_{sem}$. This latent code, analogous to those in GANs or VAEs, captures high-level semantic features of the input image.
\subsubsection{DiffMAE}
DiffMAE is an innovative method that rethinks the role of generative models in pre-training visual representations by integrating principles from denoising diffusion models. In DiffMAE, diffusion models are restructured into a masked autoencoder framework, where the input data is masked. It generates masked regions by sampling from $p(\boldsymbol{x}_0^m| \boldsymbol{x}_0^\mathbf{v})$, which is approximated by recursively sampling from $p(\boldsymbol{x}_{t-1}^m|\boldsymbol{x}_t^m,\boldsymbol{x}_0^v)$ can also considered Gaussian distributed. We optimize the simple objective proposed by DDPM:
\begin{equation}
    \mathcal{L} = \mathbb{E}_{t,\boldsymbol{x}_0,\epsilon} \| \boldsymbol{x}_0^m -D_\theta(\boldsymbol{x}_t^m,t,E_\phi(\boldsymbol{x}_0^v)) \|^2.
\end{equation}

\subsection{B.4 Downstream Backbones}
\subsubsection{DeepSurv}
Deepsurv uses deep learning to enhance the nonlinear Cox proportional hazards model through a deep linear feed-forward network. It predicts how patient covariates affect their hazard rate by learning the weights $\beta$  in the linear risk function $\hat{h}_{\beta}(\boldsymbol{x}) = \beta^T \boldsymbol{x}$. The model optimizes $\beta$ using the Cox partial likelihood:
\begin{equation}
    \mathcal{L}_{cox}(\beta) = \prod_{i:E_i=1} \frac{\text{exp}(\hat{h}_{\beta}(\boldsymbol{x}_i)}{\sum_{j \in \mathcal{R}(T_i)}\text{exp}(\hat{h}_{\beta}(\boldsymbol{x}_j))}.  
\end{equation}
Here, $T_i$,$E_i$, and $x_i$ are the respective event time, event indicator and covariates for the $i$-th observation. The risk set $\mathbb{R}(t)=\{i:T_i\ge t\}$ is the set of patients still at risk of failure at time $t$. 
\subsubsection{AB-MIL} Attention-based Deep Multiple Instance Learning proposes to use a weighted average of instances (low-dimensional embeddings) where weights are determined by a neural network. Additionally, the weights must sum to 1 to be invariant to the size of a bag. And MIL pooling as follows:
\begin{equation}
    \boldsymbol{z} = \sum_{k=1}^K \mathbf{a}_k \mathbf{h}_k
\end{equation}
where:
\begin{equation}
    \mathbf{a}_k = \frac{\text{exp}(\mathbf{w}^T \text{tanh}(\mathbf{V} \mathbf{h}_k^T)}{\sum_{j=1}^K \text{exp}(\mathbf{w}^T \text{tanh}(\mathbf{V} \mathbf{h}_j^T))}.
\end{equation}
where $\mathbf{w} \in \mathbb{R}^{L\times 1}$ and $\mathbf{V} \in \mathbb{R}^{L\times M}$ are parameters. Moreover, we utilize the $\text{tanh}(\cdot)$ element-wise non-linearity to include both negative and positive values for proper gradient flow.
\subsubsection{PatchGCN}
Patch-based Graph Convolutional Network presents a context-aware, spatially-resolved patch-based graph convolutional network that hierarchically aggregates instance-level histology features to model local and global-level topological structures in the tumor microenvironment. It makes $F_{\text{GCN}}^{(l)}$ a residual mapping and stack multiple layers of $F_{\text{GCN}}^{(l)}$ where the output of $F_{\text{GCN}}^{(l)}$ additively combines with its input. The attention pooling of instance-level features is performed in local graph neighborhoods instead of across the entire bag.
\begin{equation}
    \mathbf{G}^{(l+1)} = F_{\text{GCN}}^{(l)}(\mathbf{G}^{(l)};\bm{\phi}^{(l)},\bm{\rho}^{(l)},\bm{\zeta}^{(l)}) + \mathbf{G}^{(l)}.
\end{equation}
in which $\bm{\phi}^{(l)}$ is the additively combined node and edge features followed by ReLU activation, $\bm{\rho}^{(l)}$ is a Softmax Aggregation scheme. $\bm{\zeta}^{(l)}(\mathbf{h}_v^{(l)},\mathbf{m}_v^{(l)}) = MLP(\mathbf{h}_\mathbf{v}^{(l)}+\mathbf{m}_v^{(l)}) \rightarrow \mathbf{h}_v^{(l+1)}$.
\subsection{B.5 Evaluation Metric}
\subsubsection{Accuracy}
In the classification tasks, Accuracy measures how close a given set of observations are to their true value. Accuracy is the proportion of correctly classified instances (both positive and negative) out of the total number of instances.
\begin{equation}
    \text{Accuracy} = \frac{\text{TP}+\text{TN}}{\text{TP}+\text{TN}+\text{FP}+\text{FN}},
\end{equation}
where:
\begin{itemize}
    \item \textbf{TP}: Number of correctly predicted positive instance.
    \item \textbf{TN}: Number of correctly predicted negative instance.
    \item \textbf{FP}: Number of incorrectly predicted positive instance.
    \item \textbf{FN}: Number of incorrectly predicted negative instance.
\end{itemize}

\begin{figure*}[t!]
    \centering
    \includegraphics[width=\linewidth]{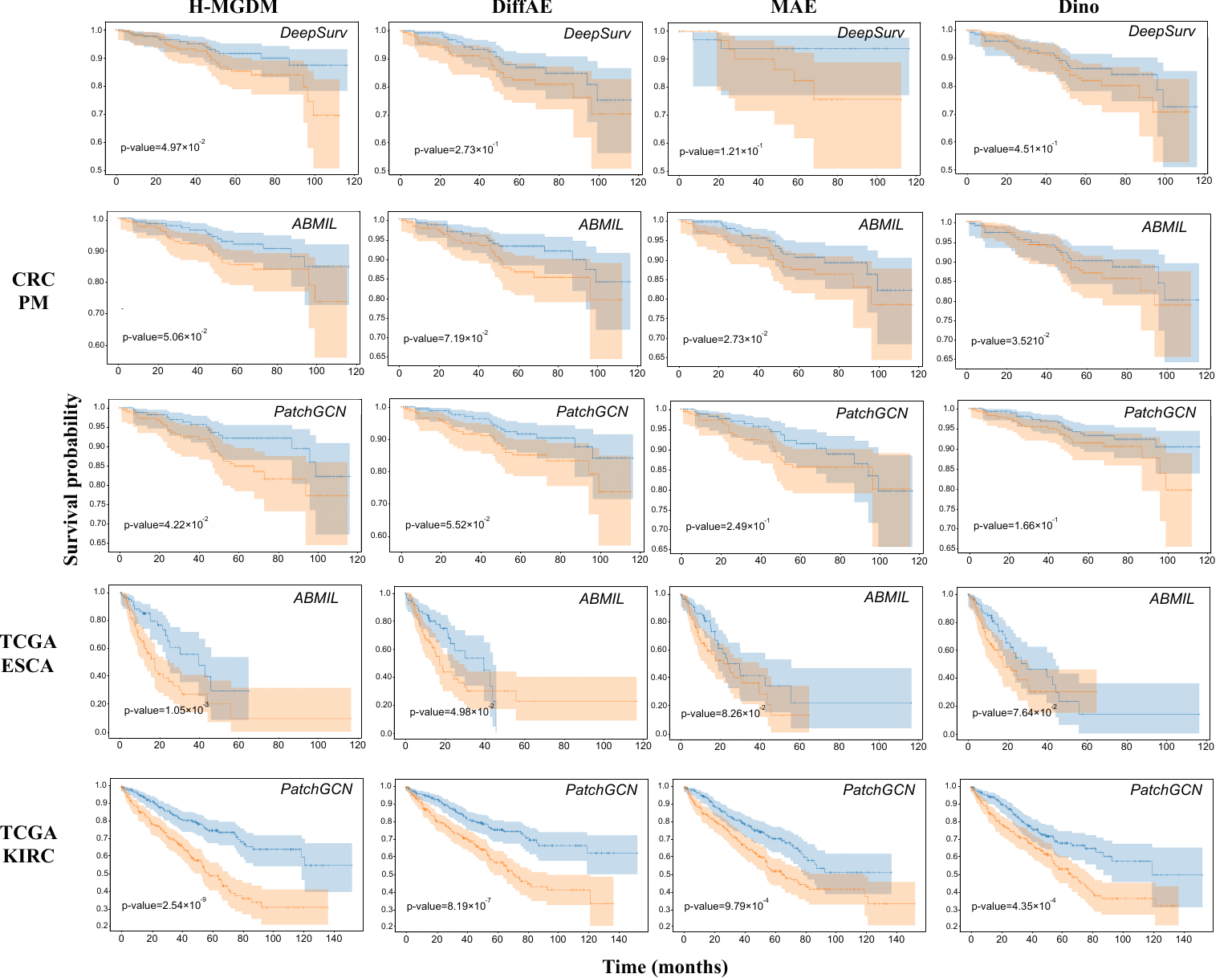}
    \caption{Kaplan-Meier Analysis of Comparison methods and our framework for three datasets on different backbones. Each cohort was split into a high-risk group (orange curve) and a low-risk group (blue curve) according to the median output of the prediction model.}
    \label{km_sup}
\end{figure*}
\begin{figure*}[t!]
    \centering
    \includegraphics[width=\linewidth]{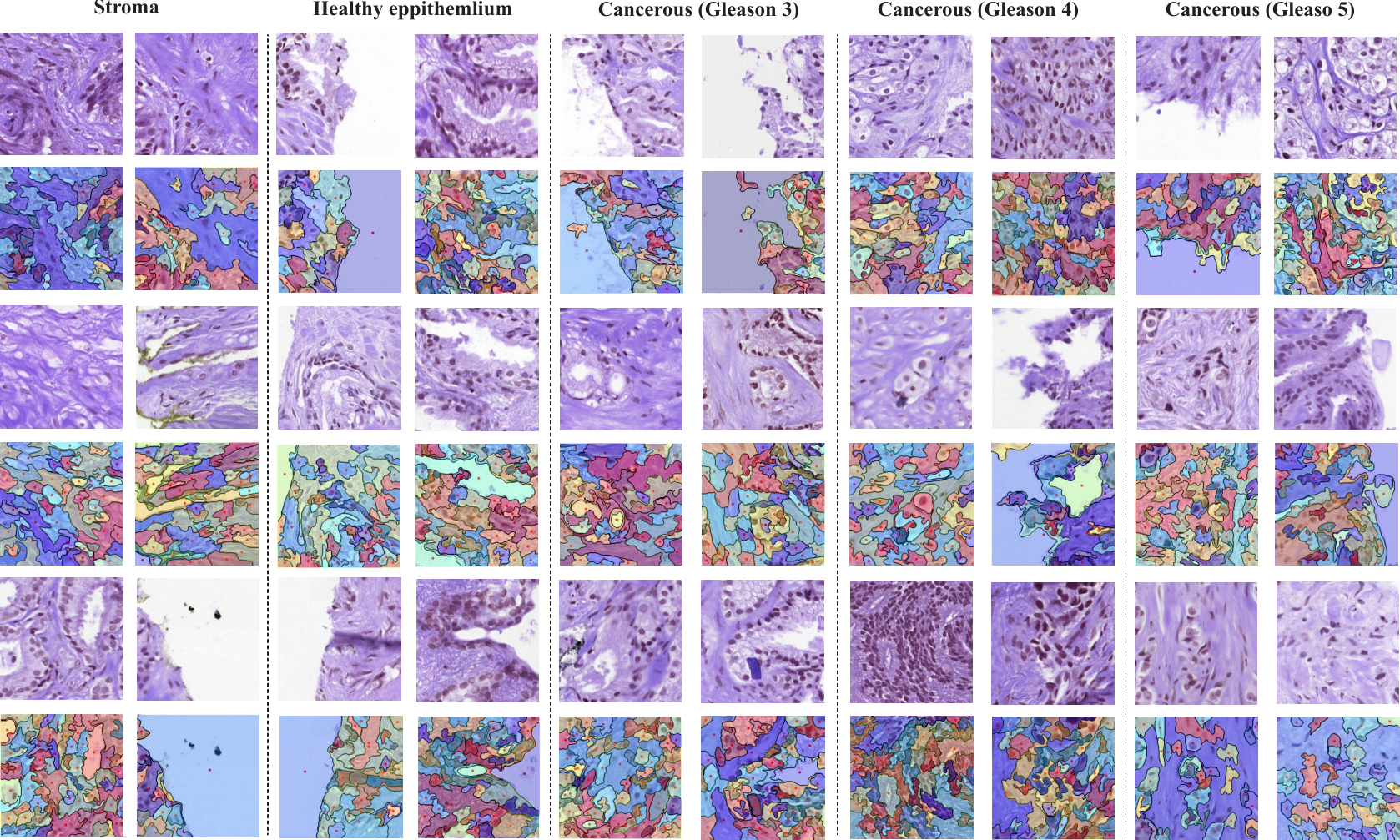}
    \caption{More Histocartographical Image Interpretability}
    \label{c2}
\end{figure*}

Accuracy measures the overall effectiveness of the model by determining the ratio of all correct predictions (both positive and negative) to the total number of cases.
\subsubsection{F1-Score}
The F1-Score is the harmonic mean of Precision and Recall, providing a single metric that balances the trade-off between these two. It is beneficial when dealing with imbalanced datasets, as it considers both false positives and false negatives.
\begin{equation}
    \text{F1} = 2 \times \frac{\text{Precision} \times \text{Recall}}{\text{Precision} + \text{Recall}}
\end{equation}
where:
\begin{itemize}
    \item \textbf{Precision} $=\frac{\text{TP}}{\text{TP+FP}
    }$: The proportion of correctly predicted positive instances out of all predicted positive instances.
    \item \textbf{Recall} $=\frac{\text{TP}}{\text{TP+FN}}$: The proportion of correctly predicted positive instances out of all actual positive instances.
\end{itemize}
\subsubsection{RMSE}
The Root Mean Square Error is a commonly used metric for evaluating the reconstruction prediction error of a model. The formula for calculating RMSE is as follow:
\begin{equation}
    \text{RMSE} = \sqrt{\frac{1}{n}\sum_{i=1}^n(y_i-\hat{y})^2},
\end{equation}
where $n$ is the number of samples. $y_i$ is the actual value of the $i$-th sample. $\hat{y}_i$ is the predicted value of the $i$-th sample. The lower the RMSE, the better the model's predictive reconstruction performace. 
\subsubsection{C-Index}
The concordance index(C-index, CI) is the most frequently used survival model evaluation metric. It is a measure of rank correlation between predicted risk scores $\hat{h}$ and observed time points $T$ with event indicator $\delta$ (1: death, recurrence, 0: not occurred). It is defined as the ratio of correctly ordered(concordant) pairs to comparable pairs. 
\begin{equation}
    \text{C-index} = \frac{\sum_{i,j}1_{T_j<T_i} \cdot 1_{\hat{h}_j > \hat{h}_i} \cdot \delta_j}{\sum_{i,j} 1_{T_j<T_i}\cdot \delta_j}.
\end{equation}
Two samples $i$ and $j$ are comparable if the sample with lower observed time $T$ experienced an event, i.e., if $T_j>T_i$ and $\delta_i=1$. A comparable pair $(i,j)$ is concordant if the estimated risk $\hat{h}$ by a survival model is higher for subjects with lower survival time, i.e., $\hat{h}_i>\hat{h}_j \wedge y_j > y_i$, otherwise the pair is discordant. 

\subsection{B.6 Experiment Environment and Dependencies}

All experiments are conducted on a workstation with 8 NVIDIA GeForce RTX 3090 (24 GB) GPUs equipped with 256 GB memory. Entity graph construction is implemented by Histocartography~\cite{jaume2021histocartography} equipped with SLIC (compactness is 10, blur kernel size is 1, threshold is 0.02). Our graph convolutional model is implemented by Pytorch Geometric \cite{Fey/Lenssen/2019}.

\section{C. Visualization and Interpretability}
\subsection{C.1 More Kaplan-Meier analysis}
We added the KM curve of the CRC-PM dataset and other backbones's result of $ESCA$ and $KIRC$ in Fig.~\ref{km_sup}.
\subsection{C.2 More Patch level Interpretability of represnetation}
More patch-level entities interpretability are shown on Fig~\ref{c2}. Our H-MGDM can more effectively focus on pathological entities related to the category and automatically ignore non-effective areas such as the background.

\section{D. Limitations and Future Work}
\subsection{D.1 Non-tile-level Entity Extraction}

For each pixel s, a window of size a × a centered at s is considered a vertex v in the pathological entity graph $P$ in pixel space. Pixels within the window that do not belong to s are assigned the background color, which can cause information redundancy and interference during superpixel feature extraction due to the excess background information. Additionally, the window size is influenced by the size of the superpixel: if the window is too small, it may not fully capture the superpixel's information, while if it is too large, it may include too much background, leaving less space for semantically meaningful parts. Therefore, a specialized feature extraction model for superpixels is urgently needed.

\subsection{D.2 Limited Amount of Data}
The data available for pretraining our model is currently limited, which poses a challenge to achieving optimal performance. Expanding the dataset with more extensive collections, especially those that encompass a broader range of cancer types, would significantly enhance both the training process and the overall effectiveness of the model. A larger and more diverse dataset would provide the model with richer information, allowing it to learn more robust patterns and generalize better across different cancer types, ultimately leading to improved predictive accuracy and reliability in clinical applications.

\end{document}